\documentclass{ecai}
\usepackage{times}
\usepackage{graphicx}
\usepackage{latexsym}
\usepackage{url}
\usepackage{textcomp}

\usepackage{times}

\usepackage{epstopdf}
\usepackage{amsmath,amssymb}

\def\vect#1{\mbox{\boldmath $#1$}}
\newcommand{\argmax}{\mathop{\rm arg~max}\limits}
\newcommand{\argmin}{\mathop{\rm arg~min}\limits}

\def\Hline{\noalign{\ifnum0=`}\fi\hrule \@height 4.\arrayrulewidth \futurelet \reserved@a\@xhline}

\begin{document}
\title{All-Transfer Learning for Deep Neural Networks and \\ its Application to Sepsis Classification}

\author{Yoshihide Sawada\institute{Advanced Research Division, Panasonic Corporation}
\and Yoshikuni Sato$^2$
\and Toru Nakada$^2$
\and Kei Ujimoto\institute{Department of Life Science, Tokyo Institute of Technology}
\and Nobuhiro Hayashi$^3$}

\maketitle
\bibliographystyle{ecai}

\begin{abstract}
In this article, we propose a transfer learning method for deep neural networks~(DNNs). 
Deep learning has been widely used in many applications. 
However, applying deep learning is problematic when a large amount of training data are not available. 
One of the conventional methods for solving this problem is transfer learning for DNNs.
In the field of image recognition, state-of-the-art transfer learning methods for DNNs
re-use parameters trained on source domain data except for the output layer.
However, this method may result in poor classification performance 
when the amount of target domain data is significantly small.
To address this problem, 
we propose a method called All-Transfer Deep Learning, which enables the transfer of all parameters of a DNN.
With this method, we can compute the relationship between the source and target labels
by the source domain knowledge.
We applied our method to 
actual two-dimensional electrophoresis image~(2-DE image) classification 
for determining if an individual suffers from sepsis;
the first attempt to apply a classification approach to 2-DE images for proteomics, which has
attracted considerable attention as an extension beyond genomics. 
The results suggest that our proposed method
outperforms conventional transfer learning methods for DNNs. 
\end{abstract}

\section{Introduction}
\label{sec:intro}
Deep learning has been widely used 
in the fields of machine learning and 
pattern recognition~\cite{bengio2009,dong2014learning,hinton2015distilling,krizhevsky2012imagenet,le2013building,zhou2014learning,zhuang2015supervised}
due to its advanced classification performance.
Deep learning is used to train a large number of parameters of 
a deep neural network~(DNN) using a large amount of training data. 
For example, Le et al.~\cite{le2013building} trained 1 billion parameters using 10 million videos, and
Krizhevsky et al.~\cite{krizhevsky2012imagenet} trained 60 million parameters using 1.2 million images.
They collected training data via the web. 
On the other hand, original data, such as biomedical data, 
cannot be easily collected due to privacy and security concerns. 
Therefore, researchers interested in solving the original task are unable to collect a sufficient amount of data to train DNNs. 
Conventional methods address this problem by applying transfer learning.

Transfer learning is a method that re-uses knowledge of the source domain 
to solve a new task of the target domain~\cite{hu2015deep,pan2011domain,pan2010survey,rosenstein2005transfer,zhuang2015supervised}.
It has been studied in various fields of AI, such as text classification~\cite{do2005transfer}, 
natural language processing~\cite{jiang2007instance}, and image recognition~\cite{saenko2010adapting}. 
Transfer learning for DNN can be divided into three approaches, supervised, semi-supervised, and unsupervised.
Recent researches focus on unsupervised domain adaptation~\cite{ganin2015unsupervised,long2015learning}.
Unsupervised and semi-supervised approach assume that the target domain labels equal to the source domain label.
However, in the biomedical field, it is difficult to collect target domain data having the same label as the source domain.
Therefore, we focus on the supervised transfer learning approach, 
which allows the labels of the source/target domain to be different.

The state-of-the-art supervised transfer learning~\cite{agrawal2014analyzing,donahue2014decaf,oquab2014learning}
construct the first~(base) model based on the source domain data by using the first cost function. 
They then construct the second model based on the target domain data 
by re-using the hidden layers of the first model as the initial values and using the second cost function.
This approach outperforms non-transfer learning when the source and target domains are similar.
However, these methods faced with the problem that causes poor classification performance and overfitting 
when the output layer has to be trained on a significantly small amount of target domain data.
Oquab et al.~\cite{oquab2014learning} and Agrawal et al.~\cite{agrawal2014analyzing}
used the Pascal Visual Object Classes~\cite{everingham2010pascal} and 
Donahue et al.~\cite{donahue2014decaf} used ImageNet~\cite{deng2009imagenet}.
The amount of target domain data of their studies was over 1,000 data points. 
On the other hand, the amount of original biomedical target domain data may be less than 100 data points.
To prevent this problem, it is necessary to re-use all layers including the output layer.
However, the method for effectively transferring the knowledge~(model) including the output layer has yet to be proposed.

In addition to the above problem, these methods are not structured to avoid negative transfer.
Negative transfer is a phenomenon that 
degrades classification accuracy when we transfer the knowledge of the source domain/task.
It is caused by using parameters computed using the data of the source domain/task irrelevant to the target task.
Although Pan et al.~\cite{pan2010survey} considered the avoidance of this phenomenon as a ``when to transfer'' problem,
little research has been published despite this important issue.
For example, Rosenstein et al.~\cite{rosenstein2005transfer}
proposed a hierarchical na\"ive Bayes to prevent this problem.
However, they did not use DNNs, and few articles have been devoted to research pertaining to DNNs.

In this article, we propose a novel method based on the transfer learning approach,
which uses two cost functions described above.
By using this approach, we can prepare the first model in advance.
It is difficult to upload the target domain data outside a hospital and
prepare a sufficient computer environment, especially for small and medium sized hospitals.
Therefore, we argue that this approach fits the clinical demand.

The main difference is that our proposed method re-uses all parameters of a DNN trained on the source domain data
and seamlessly links two cost functions by evaluating the relationship between the source and target labels
on the basis of the source domain knowledge~(Section~\ref{sec:atdl}).
By using this relationship, our method regularizes all layers including the output layer.
This means that it can reduce the risk of falling into the local-optimal solution caused by 
the randomness of the initial values.
We call our method {\it All-Transfer Deep Learning}~({\it ATDL}).

We applied ATDL to actual two-dimensional electrophoresis~(2-DE) image~\cite{rabilloud2010two,rabilloud2011two} 
classification for determining if an individual suffers from sepsis.
Sepsis is a type of disease caused by a dysregulated host response to infection 
leading to septic shock,
which affects many people around the world 
with a mortality rate of approximately 25$\%$~\cite{dellinger2013surviving,dombrovskiy2007rapid,singer2016third}.
Therefore, high recognition performance of this disease is important at clinical sites.
We use 2-DE images of proteomics to determine sepsis, 
which is currently attracting considerable attention in the biological field as the next step beyond genomics.  
In addition, we also show that there is a correlation between the relationship described above and classification performance.
This means that ATDL is possible to reduce the risk of negative transfer.
We explain 2-DE images in Section~\ref{sec:2-DE image} and explain experimental results in Section~\ref{sec:experiment}.

The contributions of this article are as follows:
\begin{itemize}
\item We propose ATDL for a significantly small amount of training data to
evaluate the relationship between the source and target labels on the basis of 
the source domain knowledge.
\item The experimental results from actual sepsis-data classification and open-image-data classifications
show that ATDL outperforms state-of-the-art transfer learning methods for DNNs,
especially when the amount of target domain data is significantly small.
\item We argue that there is a correlation between the relationship described above and classification performance.
\item This is the first attempt to apply machine learning by using DNNs to 2-DE images.
An actual sepsis-data classification accuracy of over 90\% was achieved.
\end{itemize}

\section{Two-dimensional Electrophoresis Images}
\label{sec:2-DE image}

\begin{figure}[t]
\begin{center}
\includegraphics[width=0.95\linewidth]{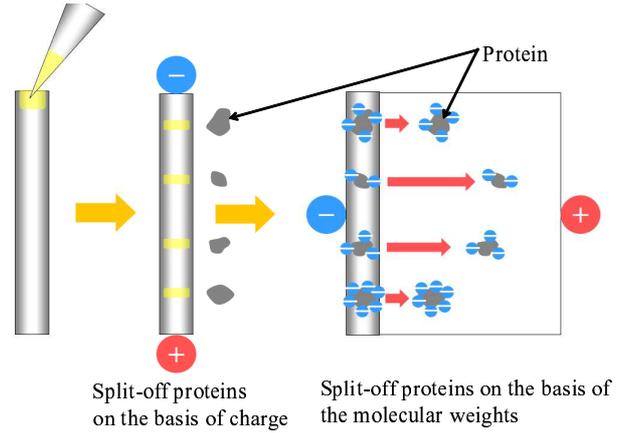}\\
\caption{Overview of production process of 2-DE images.
After extracting and refining proteins from sample,
proteins are split off by degree of isoelectric points and 
molecular weights~(SDS-PAGE~\cite{rabilloud2010two,rabilloud2011two}).}
\label{fig:2-DE image_make}
\end{center}
\end{figure}

\begin{figure}[t]
\begin{center}
\includegraphics[width=0.75\linewidth]{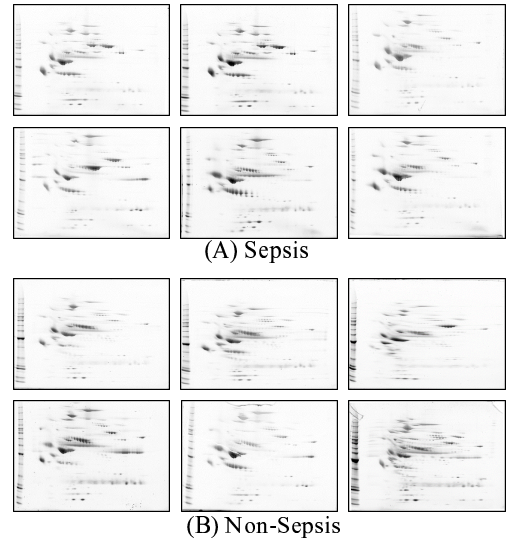}
\caption{Examples of 2-DE images.
X- and Y-axes represent degrees of molecular weights and isoelectric points, respectively, 
and black regions represent protein spots.}
\label{fig:2-DE image}
\end{center}
\end{figure}

Two-dimensional electrophoresis images represent the difference 
between the isoelectric points and molecular weights of proteins~\cite{rabilloud2010two,rabilloud2011two}.
Figure~\ref{fig:2-DE image_make} shows an overview of the process by which 2-DE images are produced, and 
Figure~\ref{fig:2-DE image} illustrates examples of 2-DE images showing sepsis and non-sepsis.
Such images are produced by first extracting and refining proteins from a sample.
After that, the proteins are split off on the basis of the degree of isoelectric points and molecular weights.
Therefore, the X-axis of 2-DE images represents the degree of molecular weights,
Y-axis represents the degree of isoelectric points,
and black regions represent the protein spots~\cite{o1975high}.

Normally, 2-DE images are analyzed for detection of a specific spot corresponding to a protein as a bio-marker, 
using computer assistance~\cite{berth2007state}.
However, many diseases, such as sepsis, are multifactorial,
which cause minute changes at many spots and unexpected spots in some cases.
Therefore, when the polymerase chain reaction~(PCR) method~\cite{bartlett2003short},
which amplifies specific genes, is applied, 
we must guess the target genes, and testing of each gene must be carried out. 
If the number of biomarkers increases, the labor will also increase. 
On the other hand, if we directly use 2-DE images, this problem can be solved because 
we can consider the comprehensive changes of proteins at one time.
From this situation, 
we try to use 2-DE images for diagnostic testing instead of using spot analysis.

\begin{figure}[t]
\begin{center}
\includegraphics[width=1.0\linewidth]{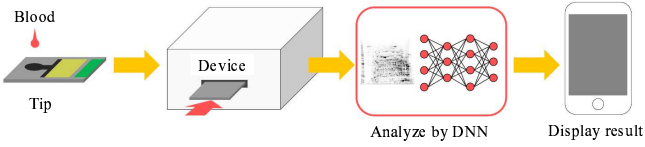}\\
\caption{Overview of our system for detecting diseases.
We focused on classification step involving analysis using DNNs.}
\label{fig:approach}
\end{center}
\end{figure}

Figure~\ref{fig:approach} shows an overview of our system for detecting diseases by using 2-DE images.
First, a doctor puts a sample of the blood of a patient
on a micro-tip, then insert it into a device that can generate 2-DE images.
Then, our system detects diseases and display the results to doctors.
The main point with our system is to detect diseases, such as sepsis,
with complex electrophoresis patterns of 2-DE images by using DNNs.
It is a matter of course that current devices for generating 2-DE images are not suitable for this concept due to issues
such as low-throughput ability and low reproducibility.
A few groups~\cite{barnes2002high,hiratsuka2007fully}
have developed techniques to generate 2-DE images
with high sensitivity, high throughput ability, and high reproducibility.
However, even if they can solve these problems in generating 2-DE images, 
collecting 2-DE images produced from patients is difficult due to privacy and security concerns. 
This clearly indicates that the need for a classification method, such as ATDL, for 
a significantly small amount of training data is increasing.

\section{All-Transfer Deep Learning}
\label{sec:atdl}

\subsection{Overview}

\begin{figure}[t]
\begin{center}
\includegraphics[width=0.95\linewidth]{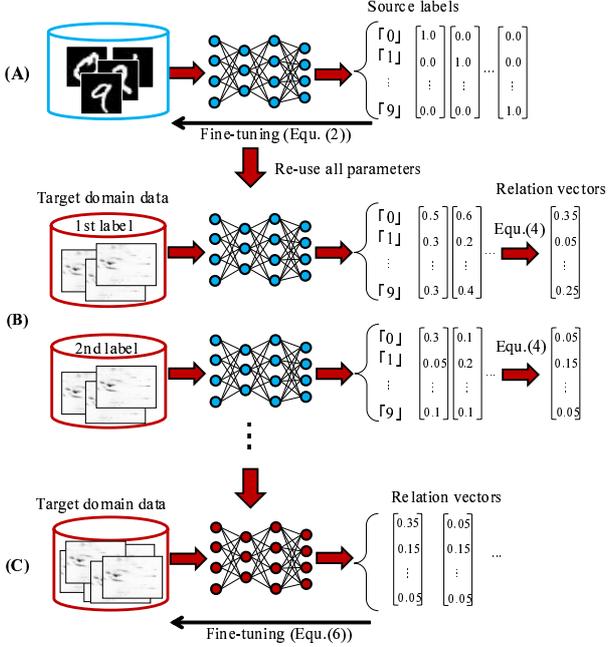}\\
\caption{Outline of ATDL. 
(A): Training DNN for source task, (B): computing relation vectors of each target label,
(C): tuning all parameters to transfer $\mathfrak{D}^s$ to $\mathfrak{D}^t$.}
\label{fig:outline}
\end{center}
\end{figure}

An outline of the ATDL training process is shown in Figure~\ref{fig:outline}.
First, ATDL trains a DNN, $\mathfrak{D}^s$, to solve the task of the source domain~(Figure~\ref{fig:outline} (A)).
In this study, we constructed $\mathfrak{D}^s$
on the basis of stacked de-noising autoencoder~(SdA)~\cite{erhan2010does,vincent2010stacked}.
Second, ATDL computes the output vectors of each target vector
by inputting them into the DNN trained on the source domain data~(Figure~\ref{fig:outline} (B)).
It then computes the {\it relation vectors} of each target label.
A relation vector denotes a vector representing the relationship between the source and target labels
on the basis of the source domain knowledge, $\mathfrak{D}^s$, by using the output vectors.
Finally, we fine-tune all parameters in such a way that the variance between the output and 
relation vectors is sufficiently small~(Figure~\ref{fig:outline} (C)).
By using the steps corresponding to Figures~\ref{fig:outline} (B) and (C), 
we can transfer $\mathfrak{D}^s$ to the DNN for the target task $\mathfrak{D}^t$,
regularizing all parameters including the output layer.
This means that ATDL provides $\mathfrak{D}^t$, which can avoid the local-optimal solution 
caused by the randomness of the initial values.

\subsection{Training Process}
\subsubsection{Construction of Deep Neural Network for Source Task}
We first explain the SdA for constructing $\mathfrak{D}^s$.
Let $\vect{x}^s \in \mathbb{R}^{D_x}$ denote a $D_x$ dimensional source vector and
$\vect{W}_i$ and $\vect{b}_i$ denote a weight matrix and bias vector of the $i$-th hidden layer~($i=1,2,\cdots,L$), respectively.
Let $\tilde{\vect{x}}$ denote a corrupting vector drawn from corruption process $q(\tilde{\vect{x}}|\vect{x})$,
and $s(.)$ denote a sigmoid function.
Then, the $i$-th hidden vector of $\mathfrak{D}^s$ is as follows,
\begin{equation}
\vect{h}_i = s(\vect{W}_i \tilde{\vect{h}}_{i-1} + \vect{b}_i).
\end{equation}
It should be noted that layer $i=0$ represents an input layer, 
that is, $\tilde{\vect{h}}_0 = \tilde{\vect{x}}^s$.
The weight and bias are computed by minimizing a de-noising reconstruction error~\cite{vincent2010stacked}.

At the output layer~(($L+1$)-th layer), we apply a regression function $f(.)$
and the cost function of $\mathfrak{D}^s$ as follows,
\begin{equation}
l( \{ \vect{y}^s, \vect{x}^s \} ) = 
\frac{1}{N^s} \sum_j^{N^s} || \vect{y}^s_j - f( \vect{h}_L | \vect{x}^s_j ) ||^2,
\label{equ:source_tuning}
\end{equation}
where $N^s$ is the amount of source domain data, 
$\vect{y}^s_j$ is a label vector 
of $\vect{x}^s_j$~($y^s_j(k) = \{ 0, 1 \}, k = 1, 2, \cdots, D_{y^s}$), 
$D_{y^s}$ is the dimension of $\vect{y}^s$, and
\begin{equation}
f( \vect{h}_L | \vect{x}^s_j ) = \vect{h}_{L+1} = \vect{W}_{L+1} \vect{h}_L + \vect{b}_{L+1}.
\end{equation}
The parameters of all layers are simultaneously fine-tuned using a stochastic gradient descent.
In this article, we use $\{ \vect{W}_i, \vect{b}_i | i = 1,2,\cdots,L+1 \}$ as the initial parameters of $\mathfrak{D}^t$.

\subsubsection{Computation of Relation Vectors}
Relation vectors represent the characteristics of the target labels
in the $D_{y^s}$ dimensional feature space computed by the source domain knowledge, $\mathfrak{D}^s$.
Let $\vect{r}_l \in \mathbb{R}^{D_{y^s}}$ denote the $l$-th relation vector~($l = 1, 2, \cdots, D_{y^t}$), 
$D_{y^t}$ denote the number of target labels,
and $\vect{x}^t_l \in \mathbb{R}^{D_x}$ denote a target vector corresponding to the $l$-th target label.
Then, $\vect{r}_l$ is computed using the following equation.
\begin{equation}
\vect{r}_l = \argmax_{\vect{h}_{L+1}} p(\vect{h}_{L+1} | \vect{x}^t_l),
\end{equation}
where $p(\vect{h}_{L+1} | \vect{x}^t_l)$ is the probability distribution of $\vect{h}_{L+1}$ given $\vect{x}^t_l$.
We assume $p(\vect{h}_{L+1} | \vect{x}^t_l)$ obeys a Gaussian distribution.
Therefore, $\vect{r}_l$ is equal to the average vector of $f(\vect{h}_L | \vect{x}^t_l)$.
\begin{equation}
\vect{r}_l = \frac{1}{N^t(l)} \sum_j^{N^t(l)} f(\vect{h}_L | \vect{x}^t_{l,j}),
\end{equation}
where $\vect{x}^t_{l,j}$ and $N^t(l)$ are the $j$-th target domain vector and
amount of target domain data corresponding to the $l$-th target label, respectively.
The $k$-th variable $r_l(k)$ means the strength of the relationship between 
the $k$-th source label and $l$-th target label.
Therefore, by confirming the values of relation vectors, 
we can understand which labels of the source domain data are similar to those of the target domain data.

\subsubsection{Fine-tuning}
After computing $\vect{r}_l$, we set $\vect{r}_l$ as the $l$-th label vector of the target task,
and all parameters including $\vect{W}_{L+1}$ and $\vect{b}_{L+1}$ are fine-tuned by 
minimizing the following main cost function using a stochastic gradient descent.
It should be noted that this equation represents the variance of the target domain data.
\begin{equation}
l( \{ \vect{r}, \vect{x}^t \} ) =
\frac{1}{N^t} \sum_l^{D_{y^t}} \sum_j^{N^t(l)} || \vect{r}_l - f(\vect{h}_L | \vect{x}^t_{l,j}) ||^2,
\label{equ:target_tuning}
\end{equation}
where $N^t = N^t(1) + N^t(2) + \cdots + N^t(D_{y^t})$.
By using this algorithm, we can have $\mathfrak{D}^t$
regularizing all parameters by using $\mathfrak{D}^s$.

\subsection{Classification Process}
In the classification process, $\mathfrak{D}^t$ 
predicts the label $\hat{l}$ of the test vector $\vect{x}$ 
on the basis of the following equation.
\begin{equation}
\hat{l} = \argmin_l (\vect{r}_l - f(\vect{h}_L | \vect{x}))^\top 
\Sigma_l (\vect{r}_l - f(\vect{h}_L | \vect{x})),
\label{equ:classify}
\end{equation}
where $\Sigma_l$ is a covariance matrix. It should be noted that classification performance does not improve
if $\vect{r}_{\hat{l}} \approx \vect{r}_{\hat{l}'}~(\hat{l} \neq \hat{l}')$.
This means that the source domain/task is not suitable for transfer.

\section{Experimental Results}
\label{sec:experiment}

We conducted experiments on 2-DE image classification for 
determining if an individual suffers from sepsis.
We compared the classification performance of five methods:
non-transfer learning, simple semi-supervised learning~(SSL), 
transfer learning by Agrawal et al.~\cite{agrawal2014analyzing}, 
that by Oquab et al.~\cite{oquab2014learning}, and ATDL.

The SSL is a method to construct a mixture model that
computes $\vect{h}_i$ using $\vect{x}^s$ and $\vect{x}^t$ and fine-tunes using only $\vect{x}^t$.
In addition, this method is a special case of that by Weston et al.~\cite{weston2012deep},
which embeds the regularizer to the output layer, when the parameter to balance between
the object function and regularizer is zero.

Agrawal's method removes the output layer of $\mathfrak{D}^s$ and adds a new output layer.
In addition to these two steps, 
Oquab's method contains an additional adaptation layer
to compensate for the different statistics of the source and target domain data.
Then, Agrawal's method fine-tunes all layers including the hidden layers~\cite{yosinski2014transferable}, and
Oquab's method fine-tunes only the adaptation and output layers.

In our study, we used a soft max function as the output layer
for constructing $\mathfrak{D}^s$ of the above transfer learning, SSL, 
and non-transfer learning methods.

To investigate the difference in classification performance,
we changed the source domain data and evaluated classification performance.
We used 2-DE images that were given different labels from the target domain data of sepsis or non-sepsis, 
MNIST~\cite{lecun1998gradient}, and CIFAR-10~\cite{krizhevsky2009learning}, as the source domain data.
In addition, to investigate the generality of our method,
we applied it to a convolutional neural network~(CNN)~\cite{lecun1998gradient}
and a different task of open-image-data classifications.
For open-image-data classifications, 
we investigated the effectiveness of our method for two open image data classifications.
Finally, we investigated the correlation coefficients 
between the classification performance and Mahalanobis distance of $\vect{r}_l$.

\subsection{Environment and Hyperparameter Settings}

\begin{table}
\begin{center}
\caption{Experimental environment.}
\begin{tabular}{c|l}
\hline
 CPU & Intel (R) Core (TM) i7-4930K\\
\hline
 Memory & 64.0GB\\
\hline
 GPU & GeForce GTX 760\\
\hline
\end{tabular}
\label{tbl:environment}
\end{center}
\end{table}

We used the computer environment summarized in Table~\ref{tbl:environment} and
pylearn2~\cite{pylearn2_arxiv_2013} to minimize (\ref{equ:source_tuning}) and (\ref{equ:target_tuning}).

In this study, we set the learning rate to $\lambda /(1.00004 \times t)$, where $t$ is the iteration.
Momentum gradually increased from 0.5 to $\mu$ when $t$ increased.
We selected the initial learning rate $\lambda$ from 
$\{ 1.0 \times 10^{-3}, 5.0 \times 10^{-3}, 1.0 \times 10^{-2}, 5.0 \times 10^{-2} \}$,
final momentum $\mu$ from $\{ 0.7, 0.99 \}$, and
size of minibatches from $\{ 10, 100 \}$.

\subsection{Actual Sepsis-Data Classification}
\label{lab:sepsis_classification}

\begin{table}
\begin{center}
\caption{List of source 2-DE images.
These images represent different extraction and refining protocols of proteins.}
\begin{tabular}{|c|l|}
\hline
 \# of source 2-DE images & Type of protocol\\
\hline
\hline
  $N^s(1)=25$ & Change amount of protein\\
\hline
  $N^s(2)=4$ & Change concentration protocol\\
\hline
  $N^s(3)=30$ & Unprocessed\\
\hline
  $N^s(4)=49$ & Removal of only top-2 abundant proteins\\
\hline
  $N^s(5)=11$ & Focus on top-2 abundant proteins\\
\hline
  $N^s(6)=15$ & Focus on 14 abundant proteins\\
\hline
  $N^s(7)=12$ & Plasma sample instead of serum\\
\hline
  $N^s(8)=19$ & Removal of Sugar chain\\
\hline
  $N^s(9)=15$ & Other protocols\\
\hline
\end{tabular}
\label{tbl:source_label}
\end{center}
\end{table}

\begin{figure}[t]
\begin{center}
\includegraphics[width=0.75\linewidth]{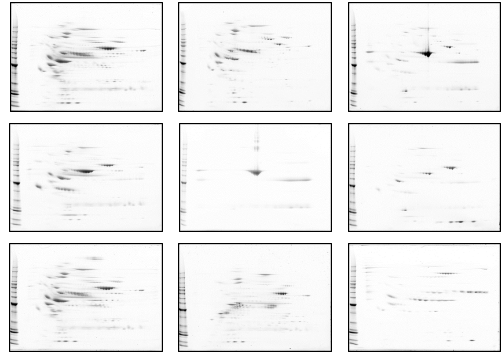}
\caption{Examples of 2-DE images that differ in extraction and refining protocol of protein.
Label number of source domain data is from 1~(top left) to 9~(bottom right) in order.}
\label{fig:source_image}
\end{center}
\end{figure}

For actual sepsis-data classification, we collected the following number of target 2-DE images $N^t = 98$,
sepsis data of $N^t(1)=30$ and non-sepsis data of $N^t(2)=68$.
The size of the 2-DE images was $53 \times 44$ pixels~($D_x=2,332$), which
was determined to save the information of the large spots.
We evaluated classification performance on the basis of two-fold cross validation.
As the source domain data, we first used 2-DE images with different labels from the target domain sepsis or non-sepsis data.
These images were generated from patients which were diagnosed as being normal.
The source task was to classify 
the differences between the extraction and refining protocols of proteins~\cite{walker2005proteomics}
shown in Table~\ref{tbl:source_label} and Figure~\ref{fig:source_image}.
As shown in this table, we set $N^s = 180$ and $D_{y^t} = 9$.
On the other hand, the target 2-DE images were generated using serum and by removing 14 abundant proteins.
These data were generated from actual patients at Juntendo University Hospital 
and were judged by infectious disease tests and SOFA/SIRS score~\cite{singer2016third}.
This study was approved by the institutional review board, and
written informed consent was obtained from patients.

\subsubsection{Comparison with Conventional Methods}

\begin{table}
\begin{center}
\caption{Classification performance of actual sepsis-data classification as function of the number of hidden layers.}
\begin{tabular}{|l|c|c|c|c|c|}
\hline
  & PPV & NPV & MCC & F1 & ACC\\
\hline \hline
 PCA + logistic regression & 0.875 & 0.805 & 0.545 & 0.609 & 0.816\\
\hline
 Non-transfer~(L=1) & 0.725 & 0.983 & 0.755 & 0.829 & 0.878\\
\hline
 Non-transfer~(L=2) & 0.718 & 0.967 & 0.726 & 0.811 & 0.867\\
\hline
 Non-transfer~(L=3) & 0.644 & 0.981 & 0.676 & 0.773 & 0.827\\
\hline
 Non-transfer~(L=4) & 0.644 & 0.981 & 0.676 & 0.773 & 0.827\\
\hline
 SSL~(L=1)          & 0.682 & \bf{1} & 0.736 & 0.811 & 0.857\\
\hline
 SSL~(L=2)          & 0.644 & 0.981 & 0.676 & 0.773 & 0.827\\
\hline
 SSL~(L=3)          & 0.592 & 0.980 & 0.620 & 0.734 & 0.786\\
\hline
 SSL~(L=4)          & 0.558 & 0.978 & 0.580 & 0.707 & 0.755\\
\hline
 Oquab et al.~\cite{oquab2014learning}~(L=1) & 0.732 & \bf{1} & 0.783 & 0.845 & 0.888\\
\hline
 Oquab et al.~\cite{oquab2014learning}~(L=2) & 0.771 & 0.952 & 0.753 & 0.831 & 0.888\\
\hline
 Oquab et al.~\cite{oquab2014learning}~(L=3) & 0.702 & 0.934 & 0.670 & 0.776 & 0.847\\
\hline
 Oquab et al.~\cite{oquab2014learning}~(L=4) & 0.658 & 0.947 & 0.648 & 0.761 & 0.827\\
\hline
 Agrawal et al.~\cite{agrawal2014analyzing}~(L=1) & 0.750 & \bf{1} & 0.800 & 0.857 & 0.898\\
\hline
 Agrawal et al.~\cite{agrawal2014analyzing}~(L=2) & 0.744 & 0.983 & 0.796 & 0.841 & 0.888\\
\hline
 Agrawal et al.~\cite{agrawal2014analyzing}~(L=3) & 0.690 & 0.982 & 0.722 & 0.806 & 0.857\\
\hline
 Agrawal et al.~\cite{agrawal2014analyzing}~(L=4) & 0.667 & \bf{1} & 0.720 & 0.8 & 0.847\\
\hline
 ATDL~(L=1) & 0.844 & 0.955 & 0.812 & 0.871 & 0.918\\
\hline
 ATDL~(L=2) & 0.871 & 0.955 & 0.834 & 0.885 & 0.929\\
\hline
 ATDL~(L=3) & 0.875 & 0.970 & \bf{0.859} & \bf{0.903} & \bf{0.939}\\
\hline
 ATDL~(L=4) & \bf{0.958} & 0.905 & 0.806 & 0.852 & 0.918\\
\hline
\end{tabular}
\label{tbl:sepsis_result}
\end{center}
\end{table}

\begin{figure}[t]
\begin{center}
\includegraphics[width=0.8\linewidth]{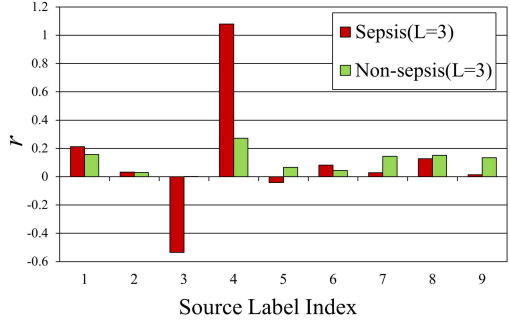}\\
\caption{Example of relation vectors of actual sepsis-data classification.}
\label{fig:sepsis_vector}
\end{center}
\end{figure}

We compared the classification performances, including that of ATDL,
with respect to the changing number of hidden layers $L = 1, 2, 3$ and $4$.
We set the dimension of the 1st hidden layer to 
$D_1=188$ by PCA using $\vect{x}^s$ and $\vect{x}^t$~(cumulative contribution of 188 features is over 99.5\%),
and $D_2$, $D_3$, and $D_4$ were set to the same dimensions.

Table~\ref{tbl:sepsis_result} lists the classification accuracies~(ACCs) of six methods
including the baseline, PCA + logistic regression~(used 188 features).
It also lists the positive predictive values~(PPVs),
negative predictive values~(NPVs), Matthews correlation coefficients~(MCCs),
and F1-scores~(F1s) as reference.
It should be noted that PPV and NPV are used in diagnostic tests,
MCC is used for evaluating performance considering the unbalance-ness of $N^t(1)$ and $N^t(2)$, 
and F1 is the harmonic value computed by precision~(=PPV) and recall.

As shown in this table, classification accuracy improved by using transfer learning.
In addition, the classification accuracy of ATDL~($L=3$) outperformed those of the other transfer learning methods.
For example, the classification accuracy of ATDL improved at least 4 percentage points
compared to that of Agrawal's method of $L=1$.
These results suggest that ATDL is effective for performing actual sepsis-data classification.

Figure~\ref{fig:sepsis_vector} shows an example of relation vectors of sepsis and non-sepsis~($L=3$).
The red bars represent the relation vector of sepsis, whereas the green bars represent that of non-sepsis.
The numbers on the X-axis correspond to the source label indices listed in Table~\ref{tbl:source_label}.
As shown in this figure, the relation vectors of sepsis and non-sepsis differed.
These results suggest that $\vect{r}_l$ can represent the characteristics of the target label 
in the feature space computed by $\mathfrak{D}^s$.

\subsubsection{Comparison of Various Source Tasks}
\label{subsub:task}

\begin{table}
\begin{center}
\caption{Classification performance of actual sepsis-data classification for different source tasks.}
\begin{tabular}{|l|c|c|c|c|c|}
\hline
  & PPV & NPV & MCC & F1 & ACC\\
\hline \hline
 Non-transfer~($D_i=188$) & 0.718 & 0.967 & 0.726 & 0.811 & 0.867\\
\hline
 Non-transfer~($D_i=500$) & 0.644 & 0.981 & 0.676 & 0.773 & 0.827\\
\hline
 Non-transfer~($D_i=1,000$) & 0.7 & 0.966 & 0.709 & 0.8 & 0.857\\
\hline
 CIFAR-10~($D_i=188$) & 0.657 & 0.889 & 0.568 & 0.708 & 0.806\\
\hline
 CIFAR-10~($D_i=500$) & \bf{0.923} & 0.912 & 0.804 & 0.857 & 0.918\\
\hline
 CIFAR-10~($D_i=1,000$) & 0.690 & \bf{0.982} & 0.722 & 0.806 & 0.857\\
\hline
 MNIST~($D_i=188$) & 0.778 & 0.968 & 0.780 & 0.849 & 0.898\\
\hline
 MNIST~($D_i=500$) & 0.839 & 0.940 & 0.786 & 0.852 & 0.908\\
\hline
 MNIST~($D_i=1,000$) & 0.828 & 0.913 & 0.735 & 0.813 & 0.888\\
\hline
 2-DE image~($D_i=188$) & 0.875 & 0.970 & \bf{0.859} & \bf{0.903} & \bf{0.939}\\
\hline
 2-DE image~($D_i=500$) & 0.844 & 0.955 & 0.812 & 0.871 & 0.918\\
\hline
 2-DE image~($D_i=1,000$) & 0.824 & 0.969 & 0.818 & 0.875 & 0.918\\
\hline
\end{tabular}
\label{tbl:sepsis_result_dif_source}
\end{center}
\end{table}

\begin{figure}[t]
\begin{center}
\includegraphics[width=0.8\linewidth]{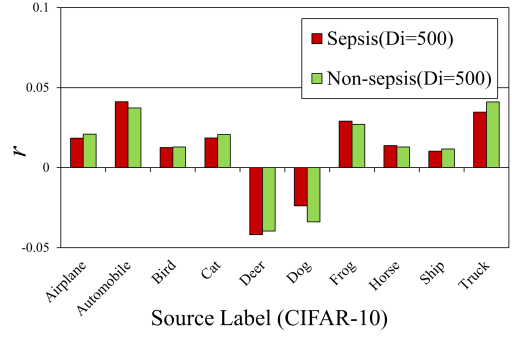}\\
\caption{Example of relation vectors when source domain data are from CIFAR-10.}
\label{fig:cifar_vector}
\end{center}
\end{figure}

\begin{figure}[t]
\begin{center}
\includegraphics[width=0.8\linewidth]{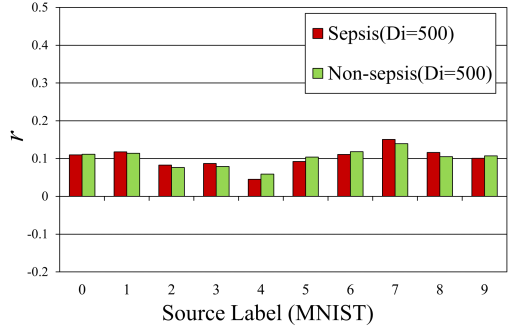}\\
\caption{Example of relation vectors when source domain data are from MNIST.}
\label{fig:mnist_vector}
\end{center}
\end{figure}

We compared classification performance with respect to changing the source domain data, which were obtained from
MNIST, CIFAR-10, and 2-DE images.
The number of images extracted from MNIST and CIFAR-10 were $N^s = 50,000$.
The CIFAR-10 images were converted to gray-scale, and
the MNIST and CIFAR-10 images were resized to $D_x= 53 \times 44 = 2,332$ to ensure they were aligned with the 2-DE images.
In addition, we set $L=3$ and $D_1=D_2=D_3$.
We also evaluated classification performance
with respect to changing the dimension of each hidden layer~$D_i = 188, 500,$ and $1,000$.

Table~\ref{tbl:sepsis_result_dif_source} lists the classification accuracies, and
the PPVs, NPVs, MCCs, and F1s as reference.
The classification accuracy based on the use of 2-DE images as the source domain data 
was higher than those obtained with MNIST and CIFAR-10,
although the number of 2-DE images was smaller~($N^s=180$).

Figure~\ref{fig:cifar_vector} shows an example of the relation vectors using CIFAR-10~($D_i=500$) and 
Figure~\ref{fig:mnist_vector} shows them using MNIST~($D_i=500$).
Compared to Figure~\ref{fig:sepsis_vector}, 
the relation vector of sepsis was considerably closer to that of non-sepsis.

These results show that information on the differences between the extraction and refining protocols of proteins
is useful for classifying sepsis, rather than using CIFAR-10 and MNIST.
Namely, if we collect the source domain data,
we have to consider the relationship between the source and target domain data.

\subsubsection{Applying ATDL to Convolutional Neural Network}

\begin{figure}[t]
\begin{center}
\includegraphics[width=0.8\linewidth]{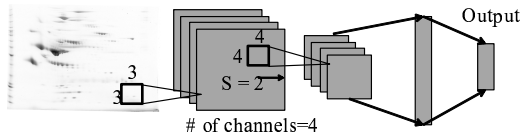}\\
\caption{CNN structure.}
\label{fig:cnn_structure}
\end{center}
\end{figure}

\begin{figure}[t]
\begin{center}
\includegraphics[width=0.8\linewidth]{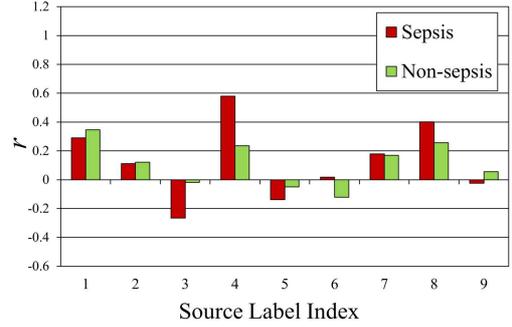}\\
\caption{Relation vectors when ATDL was applied to CNN.
Numbers on X-axis correspond to source label indices.}
\label{fig:cnn_vector}
\end{center}
\end{figure}

\begin{table}
\begin{center}
\caption{Classification performance when ATDL was applied to CNN.}
\begin{tabular}{|l|c|c|c|c|c|}
\hline
 & PPV & NPV & MCC & F1 & ACC\\
\hline \hline
 Non-transfer & 0.717 & 0.966 & 0.726 & 0.812 & 0.867\\
\hline
 ATDL & \bf{0.829} & \bf{0.984} & \bf{0.845} & \bf{0.892} & \bf{0.929}\\
\hline
\end{tabular}
\label{tbl:sepsis_result_cnn}
\end{center}
\end{table}

To investigate the effectiveness of our method regarding other DNNs, we applied it to a CNN
and evaluated its classification performance by using 2-DE images as the source domain data.
Figure~\ref{fig:cnn_structure} shows the structure of the CNN, which
was determined on the basis of two-fold cross validation with respect to changing the hyperparameters shown in this figure.

Table~\ref{tbl:sepsis_result_cnn} lists the classification performances.
The ATDL performed better than the non-transfer learning method and 
approximately equal to the SdA of ATDL~($L=3$) shown in Table~\ref{tbl:sepsis_result}.
Thus, these results suggest that ATDL is applicable to CNNs as well as SdAs.
The CNN is widely used in image recognition and 
achieves high classification accuracy on several standard data~\cite{jia2014caffe,krizhevsky2012,le2013building}.
Therefore, we consider that 
ATDL is possible to be applied to various image recognition problems.

Figure~\ref{fig:cnn_vector} shows the relation vectors.
Sepsis had a relationship to the 4th source label~(removal of only top-2 abundant proteins),
which is the same as in Figure~\ref{fig:sepsis_vector}.
This result suggests that there are biological relationships between them.
In the future, we plan to examine this result from a biological point of view.

\subsection{Open-Image Data Experiment}

\begin{table}
\begin{center}
\caption{Accuracy of automobile and pedestrian crossing.}
\begin{tabular}{|l|c|c|}
\hline
 \# of target images $N^t$ & 400 & 1,500 \\
\hline \hline
 Non-transfer & 0.724 & 0.753 \\
\hline
 SSL & 0.750 & 0.789 \\
\hline
 Oquab et al.~\cite{oquab2014learning} & 0.763 & 0.782\\
\hline
 Agrawal et al.~\cite{agrawal2014analyzing} & 0.753 & 0.781\\
\hline
 ATDL & \bf{0.789} & \bf{0.797}\\
\hline
\end{tabular}
\label{tbl:simu_result}
\end{center}
\end{table}

\begin{table}
\begin{center}
\caption{Accuracy of MNIST.}
\begin{tabular}{|l|c|c|c|}
\hline
 \# of target images $N^t$  & 1,000 & 5,000 & 10,000 \\
\hline \hline
 Non-transfer & 0.854 & 0.926 & 0.945\\
\hline
 SSL & 0.844 & \bf{0.928} & \bf{0.951}\\
\hline
 Oquab et al.~\cite{oquab2014learning} & 0.773 & 0.875 & 0.887\\
\hline
 Agrawal et al.~\cite{agrawal2014analyzing} & 0.844 & 0.923 & \bf{0.951}\\
\hline
 ATDL & \bf{0.887} & \bf{0.928} & 0.932 \\
\hline
\end{tabular}
\label{tbl:simu_result2}
\end{center}
\end{table}

\begin{figure}[t]
\begin{center}
\includegraphics[width=0.8\linewidth]{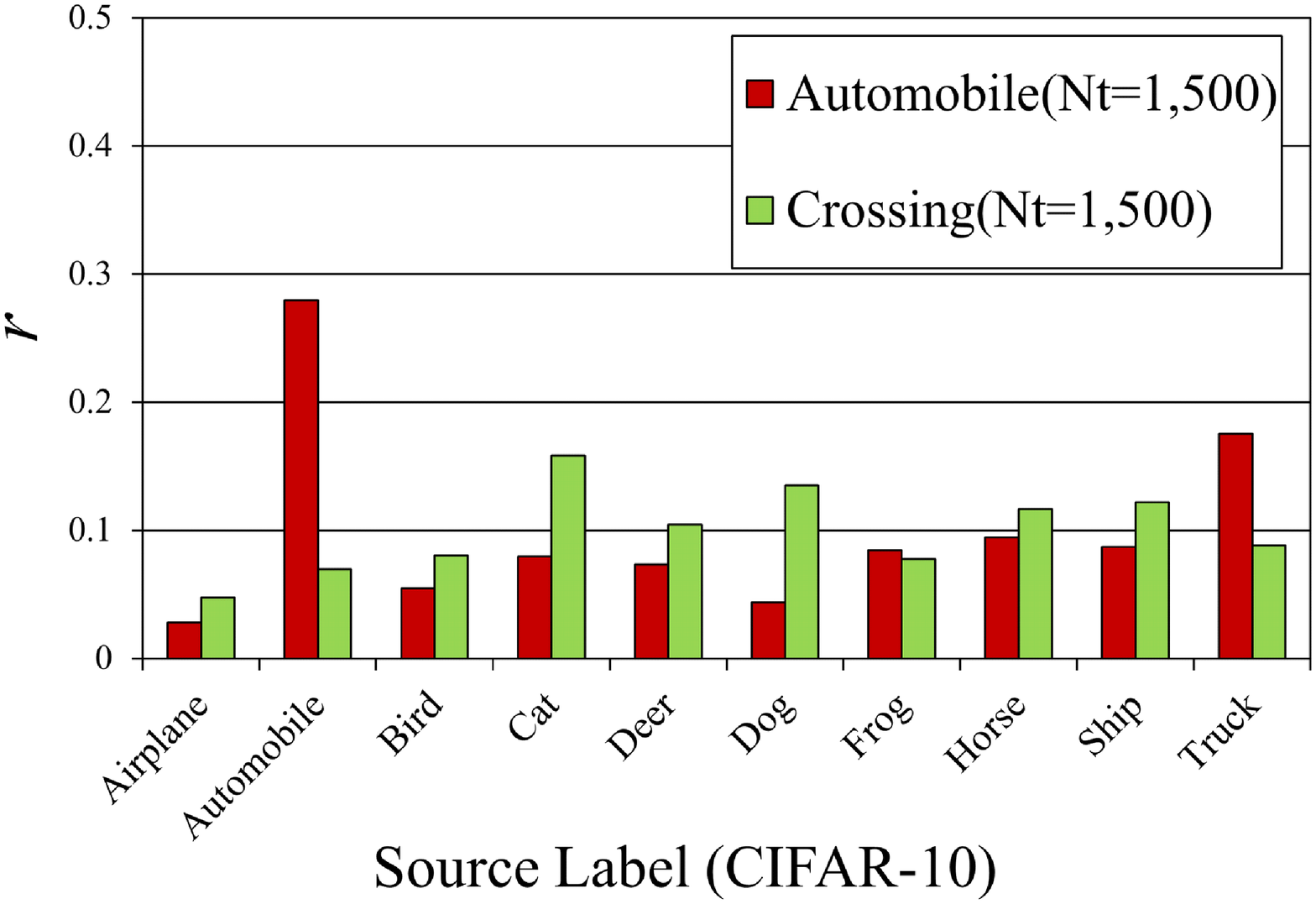}\\
\caption{Example of relation vectors of task (1).}
\label{fig:simu_vector}
\end{center}
\end{figure}

\begin{figure}[t]
\begin{center}
\includegraphics[width=0.8\linewidth]{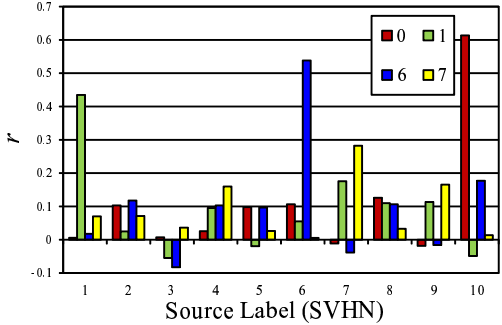}\\
\caption{Example of relation vectors of task (2)~($N^t=1,000$).}
\label{fig:simu_vector2}
\end{center}
\end{figure}

To investigate the generalization of our method,
we first applied our method to two open image data classifications:
(1) CIFAR-10~\cite{krizhevsky2009learning} as the source domain and
images of an automobile and pedestrian crossing from ImageNet~\cite{krizhevsky2012imagenet} as the target domain,
and (2) SVHN~\cite{netzer2011reading} as the source domain and MNIST~\cite{lecun1998gradient} as the target domain.
Task (1) is an example in which the source/target domain data consist of color images, 
and (2) is an example of multiclass classification.

For task (1), we constructed $\mathfrak{D}^s$ on the basis of the SdA and
set $L=3$, $D_{y^s}=10$, $D_i=1,000~(i=1,2,3)$, and $N^s=50,000$.
As the target test data, we used $750$ images of automobile and $750$ images of pedestrian crossing.
For task (2), we also constructed $\mathfrak{D}^s$ on the basis of the SdA and
set $L=3$, $D_{y^s}=10$, $D_i=100~(i=1,2,3)$, $N^s=73,257$, and SVHN images were converted to gray-scale.
As the target test data, we used $10,000$ images from MNIST.
Test images of task (1) and (2) were not included in the training target domain data.

Table~\ref{tbl:simu_result} and \ref{tbl:simu_result2} list the classification accuracies of the five methods 
for different amounts of target domain data.
It should be noted that we could not conduct actual sepsis-data classification 
in this experiment because collecting sepsis data is difficult.
As shown in these tables, our method outperformed other methods when 
$N^t = 400, 1,500$ for task (1) and $N^t = 1,000$ for task (2).
These results suggest that our method is effective when
the amount of target domain data is significantly small.

Figure~\ref{fig:simu_vector} and \ref{fig:simu_vector2} show examples of the relation vectors of each task.
As shown in these figures, 
the target automobile showed a relationship with the source automobile~(2nd source label) and source truck~(10th source label).
In addition, the highest relation of the character ``6'' of MNIST was the character ``6'' of SVHN.
On the other hand, the relation vectors of the target automobile/pedestrian crossing differed.
These results suggest that $\vect{r}_l$ enabled the representation of the target label characteristics
in the feature space computed by $\mathfrak{D}^s$, the same as 2-DE images.

\subsection{Correlation of Performance and Distance}

\begin{figure}[t]
\begin{center}
\includegraphics[width=0.95\linewidth]{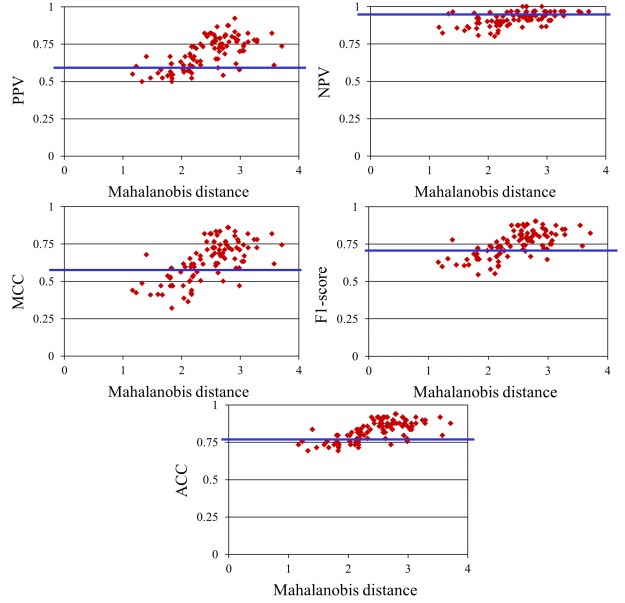}
\caption{Relationship between Mahalanobis distance and classification performance.
Blue lines represent classification performance of non-transfer learning method.}
\label{fig:correlatoin}
\end{center}
\end{figure}

\begin{table}
\begin{center}
\caption{Correlation coefficient $R$ and p-value 
of each classification performance.}
\begin{tabular}{|l|c|c|c|c|c|}
\hline
         & PPV & NPV & MCC & F1 & ACC \\
\hline \hline
 $R$     & 0.627 & 0.521 & 0.663 & 0.657 & 0.665 \\
\hline
 p-value & $< 0.01$ & $< 0.01$ & $< 0.01$ & $< 0.01$ & $< 0.01$\\
\hline
\end{tabular}
\label{tbl:correlation}
\end{center}
\end{table}

As described above, the classification performance of ATDL depends on the distance of $\vect{r}_l$.
In this subsection, we discuss the investigation of the correlation between classification performance and 
Mahalanobis distance $d_m$.
If they correlate, ATDL can be used to select $\mathfrak{D}^s$ before the fine-tuning process
and reduce the risk of negative transfer.

Let $M$ denote the number of source domain sub-groups,
$N^s_a~(a=1,2,\cdots,M)$ denote the amount of $a$-th source domain sub-group data, 
and $\vect{X}_a = \{ \vect{y}^s_b, \vect{x}^s_b | b = 1,2,\cdots, N^s_a \}$ 
denote the $a$-th source domain sub-group data sampled from all source domain data $\vect{X}$.
We constructed $\mathfrak{D}^s_a$ by using $\vect{X}_a$ and computed the Mahalanobis distance $d_m(a)$
of $\vect{r}_l$ by inputting the $a$-th DNN $\mathfrak{D}^s_a$.
Then, we fine-tuned to transfer from $\mathfrak{D}^s_a$ to $\mathfrak{D}^t$.

We used MNIST as $\vect{X}$ and set $M=100$ and $N^s_a = 5,000~(a=1,2,\cdots,M)$.
Sub-groups were randomly selected from $\vect{X}$, and
$\mathfrak{D}^s_a$ was constructed on the basis of the SdA.
The target task was sepsis-data classification, and we set $L=1$, $D_1=188$, and $N^t=49$~($N^t(1)=15, N^t(2)=34$).
As the target test data, we used $15$ images of sepsis and $34$ images of non-sepsis.
It should be noted that all hyperparameters were fixed.

To evaluate the relationship between $d_m(a)$ and classification performance $t(a)$, 
we computed the correlation coefficient $R$ as follows.
\begin{equation}
R = \frac{\sum_a^M (d_m(a)-\bar{d}_m)(t(a)-\bar{t})}{\sqrt{\sum_a^M(d_m(a)-\bar{d}_m)^2 \sum_a^M(t(a)-\bar{t})^2}},
\end{equation}
where $\bar{d}_m$ and $\bar{t}$ are the averages of $d_m$ and $t$.
Figure~\ref{fig:correlatoin} shows $d_m(a)$ and the corresponding classification performances,
and Table~\ref{tbl:correlation} lists the correlation coefficients and p-values.
The blue lines in Figure~\ref{fig:correlatoin} represent the performance of non-transfer learning.
The Mahalanobis distance and classification performances correlated, suggesting that
higher classification performance than that of non-transfer learning
is possible by using $\mathfrak{D}^s_a$ with large $d_m(a)$.
This means that we can select $\mathfrak{D}^s_a$ effectively before the fine-tuning process.

\section{Conclusion}
We proposed ATDL, a novel transfer learning method for DNNs, 
for a significantly small amount of training data. 
It computes the relation vectors that represent the characteristics of target labels by the source domain knowledge.
By using the relation vectors, ATDL enables the transfer of all knowledge of DNNs including the output layer.

We applied ATDL to actual sepsis-data classification.
The experimental results showed that ATDL outperformed other methods.
We also investigated the generality of ATDL with respect to changing the DNN model and target task, and
compared the classification performance with respect to changing the source domain data.
From the results, we argue that our method is applicable to other tasks, especially when
the amount of target domain data was significantly small, and
classification performance improves when 
we use the source domain data that are similar to the target domain data.
Furthermore, we showed the possibility of selecting an effective DNN before the fine-tuning process.

To the best of our knowledge, this work involved the first trial 
in which 2-DE images were analyzed using the classification approach, which resulted in over 90\% accuracy.
Thus, this article will be influential not only in machine learning, but also medical and biological fields.

In the future, we will collect source domain 2-DE images that can be uploaded easily, apply ATDL to a deeper network,
predict classification performance more accurate, 
and analyze the relation vectors from a biological point of view.

\section{Acknowledgments}
We would like to acknowledge Dr. Iba, Professor of Juntendo University School of Medicine,
for his help in collecting samples to generate 2-DE images.

\bibliography{myref}
\end{document}